# A Knowledge-Oriented Approach to Enhance Integration and Communicability in the Polkadot Ecosystem

Marcio Ferreira Moreno
MOBR Systems
Sao Paulo, Brazil
marcio@mobr.ai

Rafael Rossi de Mello Brandão
MOBR Systems
Sao Paulo, Brazil
rafael@mobr.ai

*Abstract* — The Polkadot ecosystem is a disruptive and highly complex multi-chain architecture that poses challenges in terms of data analysis and communicability. Currently, there is a lack of standardized and holistic approaches to retrieve and analyze data across parachains and applications, making it difficult for general users and developers to access ecosystem data consistently. This paper proposes a conceptual framework that includes a domain ontology called POnto (a Polkadot Ontology) to address these challenges. POnto provides a structured representation of the ecosystem's concepts and relationships, enabling a formal understanding of the platform. The proposed knowledge-oriented approach enhances integration and communicability, enabling a wider range of users to participate in the ecosystem and facilitating the development of AI-based applications. The paper presents a case study methodology to validate the proposed framework, which includes expert feedback and insights from the Polkadot community. The POnto ontology and the roadmap for a query engine based on a Controlled Natural Language and POnto, provide valuable contributions to the growth and adoption of the Polkadot ecosystem in heterogeneous socio-technical environments.

## I.   Introduction

The Polkadot ecosystem [10] along with its multi-chain architecture presents a unique set of challenges in terms of data analysis and communicability. Polkadot is a highly disruptive ecosystem, however it is relatively new and still lacks a standardized and holistic approach to retrieve and analyze data across parachains and applications. This makes it challenging for general users to access ecosystem data and for developers to write consistent and efficient querying code that works across the entire network.

Currently, Polkadot literature is predominantly written either from a purely technical or from an economical perspective. The main goal of this paper is to support the growth of the Polkadot community in both formal knowledge and applications, since it is critical that users, developers, and stakeholders have a formal and fundamental understanding on how the platform is built and how it works. Without such formal representation, application developers may struggle to integrate software components and data, as well as to build querying and reasoning features in a standardized manner. Unstructured development combined with lack of expressiveness about the ecosystem may jeopardize interoperation and wide adoption of Polkadot, especially in heterogeneous socio-technical environments.

The application of knowledge-oriented techniques to blockchain is not a novelty per se. Different initiatives [1, 2, 3, 4, 5] explored various perspectives and approaches towards structuring, applying and reusing domain knowledge in Web3. In addition, initiatives like the ones discussed in [6] focus on the use of ontologies for enhancing communication and integration in communities. However, none of these works present contributions to structure the retrieval, integration, and analysis of data from different sources within blockchain ecosystems.

On top of that, Polkadot is a technology that is constantly growing and evolving, as such there is a need for effective mechanisms to enhance the exchange of domain knowledge among its participants. In response to this need, we present a conceptual framework that includes a domain ontology called POnto (a Polkadot Ontology). The conceptual framework comprises a structured and organized system of concepts, principles, and assumptions that form the foundation of how symbolic artificial intelligence systems are designed and developed. This framework helps in understanding and representing knowledge, reasoning, and problem-solving processes using symbols and rules.

The Polkadot community and the ecosystem as whole are expected to benefit from the proposed conceptual framework, the main contribution of this work. By enhancing integration and communicability through a knowledge-oriented approach, the framework enables a wider range of users to participate in the ecosystem and facilitates the development of AI-based applications. The proposed approach is validated through a case study methodology, which includes pilot testing, data collection, analysis, reporting findings, and gathering expert feedback.

The remainder of this paper is organized as follows. Section II discusses the background that underlies this work. Section III presents the main related work. Section IV discusses the methodology applied to accomplish this work. Section V introduces the POnto ontology, while Section VI discusses its validation. Finally, Section VII presents the final remarks and points out future work.

## II. BACKGROUND

In this section, we discuss existent alternatives to enhance blockchain technologies using ontologies. The discussed alternatives base the purpose and design of the proposed conceptual framework. Cano-Benito et al. [1] discuss six scenarios for integrating ontologies and blockchains. Fig. 1 illustrates the first scenario of a chain of blocks with meta-data expressed following an ontology, and the content of the blocks is expressed in a non-RDF format.

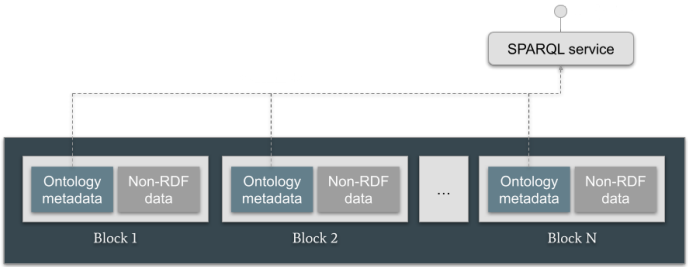

Fig. 1. Blockchain with metadata referencing ontology.

Fig. 2 illustrates a variation of the first scenario where content of the blocks may be expressed in any format that RDF supports, e.g. Turtle, JSON-LD or XML/RDF.

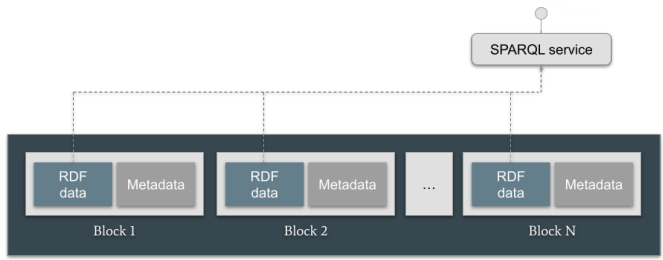

Fig. 2. Blockchain with RDF content.

Virtual RDF services such the one illustrated in Fig. 3 can be used for linking data and combining several data on-the-fly. They consume a data source (e.g, the blockchain or other data structure) and output RDF. Services can publish the data as a dataset and provide a SPARQL query endpoint, or only generate an RDF dump to be stored in a triple store in order to query the data.

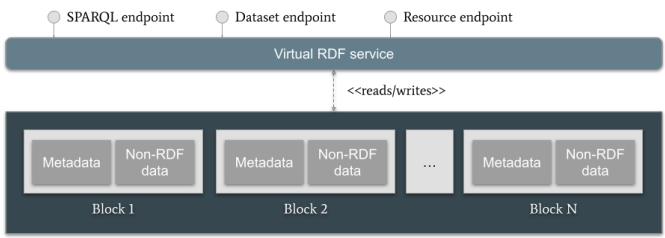

Fig. 3. Blockchain and Virtual RDF.

According to the authors, the possible implementation depicted in Fig. 4 would benefit the semantic web stack, since a blockchain could be used to maintain immutable and trustworthy references to external RDF datasets. In a way, in this scenario the blockchain could act as a DNS to identify RDF fragments, even in case DNS servers change and no longer identify RDF elements URI.

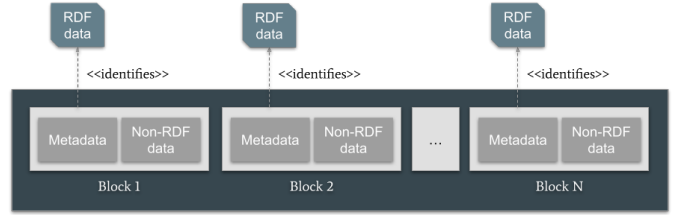

Fig. 4. Blockchain with external pointers.

Fig. 5 shows a variation with cross-chain references. A first chain is used to identify RDF resources stored in a second chain implemented through any of the previous approaches. RDF data would be immutable, transparent, and double identified (by the URIs and the hashes of the first blockchain).

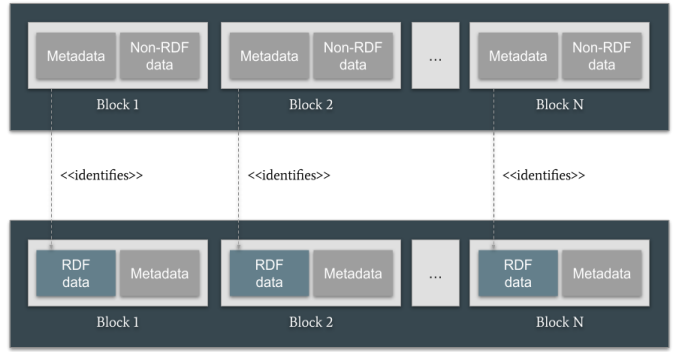

Fig. 5. Blockchain referencing another blockchain.

The scenario illustrated in Fig. 6 consists of a blockchain implementation which would take advantage of all the semantic web technologies from scratch. That is, a knowledge-oriented blockchain implementation. No such implementation is currently available.

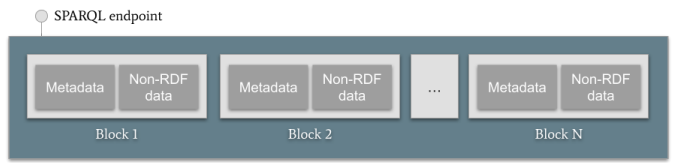

Fig. 6. Blockchain implementation relying on Semantic Web.

The purpose of the POnto ontology is to structure Polkadot ecosystem's concepts and relationships, enabling query ability for correlating data from existing blockchains (i.e., relay chains, and parachains) in the ecosystem. Such ability demands a service capable of reading and structuring chain data, blocks, transactions, smart contracts, etc., on the fly into knowledge graphs. In this sense, a service similar to the depicted in Fig. 3 would be more appropriate for such requirements.

## III. RELATED WORK

Among the related work, we considered studies investigating the relation and impact of semantic tools on Web3 projects regarding data integration and communicability. In addition, we observed well-established knowledge-oriented studies for Web3 available in the literature.

Cano-Benito et al. [1] discuss the sharing and reusing of ontologies in a peer-to-peer community, highlighting the potential benefits of ontologies in the blockchain domain.

Velasco [2] examines the political perspective of blockchain ontologies and distributed ledger technologies, discussing their potential impact on various domains, including open science and academic publishing. Janowicz et al. [3] explore the prospects of blockchain and distributed ledger technologies for open science and academic publishing, highlighting their potential benefits and challenges.

Scrocca et al. [4] propose a framework that combines semantic web technologies and smart contracts for efficient contract management in the transportation industry, demonstrating the effectiveness of this approach through a case study. Ding et al. [5] present DeSci, a comprehensive overview and reference model based on Web3 and DAO, which incorporates semantic web technologies to enhance decentralized scientific collaboration.

Another work by Palma et al. [6] focuses on the use of ontologies in the World Wide Web, emphasizing the importance of semantic web technologies for enhancing communication and integration.

Specifically considering ontologies applied to Web3, the Ethon Ontology [7] is one of the most disseminated and well-established. It is a formal representation of the Ethereum ecosystem that aims to provide a standardized vocabulary and ontology for Ethereum-related concepts and entities. It is modeled using Web Ontology Language (OWL) and is closely aligned with the Ethereum yellow paper. Use cases include allowing semantically annotating content provided by Ethereum based tools and DApps. EthOn is organized into several modules, each of which corresponds to a specific aspect of the Ethereum ecosystem. These modules include: Core (fundamental classes and properties), Blockchain (blocks, transactions, and mining, etc.), Contracts (smart contracts, their state variables, and functions), Tokens (including ERC20, ERC721, and ERC1155 tokens), and DeFi (decentralized exchanges, liquidity pools, lending protocols, and others).

BLONDIE (BLockchain ONtology with Dynamic Extensibility) [8] is an ontology expressed in OWL language aiming to support interoperability among different blockchain systems, in its current version: Bitcoin, Ethereum and IBM Hyperledger. It provides a common vocabulary for describing chain-specific concepts and relationships. This can be useful for developing applications that need to interact with multiple blockchains or for comparing and contrasting different blockchain systems.

A potential limitation of the ontology is that it may not capture all the nuances and details of other blockchain systems, as it does not aim to provide a general and abstract model of blockchain technologies. Nonetheless, BLONDIE ontology represents an important contribution to the development of a standardized vocabulary for blockchain technology and has the potential to facilitate the development of interoperable blockchain applications.

The DLT (Distributed Ledger Technology) Ontology [9] defines a set of concepts and relationships that capture the essential characteristics of DLT systems, including their architecture, consensus mechanisms, transaction models, and data structures. The ontology is designed to be modular and extensible, allowing developers and researchers to tailor it to their specific needs. It includes concepts to model security aspects such as technical threats and vulnerabilities of distributed ledger systems, application domains, as well as relevant standards and regulations. The ontology was developed based on collection of information and competency questions. It covers technical setup and components of DLT systems, security aspects, as well as applications and use cases. It covers 115 classes and 15 properties, and consists of a total of 571 triples.

The related work highlights the significance of ontologies and semantic web technologies to improve communication, integration, and structured development in the blockchain domain. The proposed knowledge-oriented approach builds upon these ideas to address related challenges in the Polkadot ecosystem [4].

To the best of our knowledge, none of the works observed in the literature address the challenge of creating a conceptual framework for multi-chain ecosystems, capable of correlating domain knowledge, targeting support for holistic analyses across multiple chains.

## IV. METHODOLOGY

To guide the design of the conceptual framework, we proposed a research strategy based on a case study approach [11]. We conducted an investigation following a protocol targeting a transparent methodology to ensure reproducibility and minimize bias. It comprised eight steps including, defining the study objectives, defining and selecting participants, developing a questionnaire to gather information, pilot-testing the questionnaire, collecting data through surveys and online sessions, analyzing the gathered data, summarizing findings and gathering feedback over achieved results.

The objectives of the study were twofold. First, to assess the perceived value and effectiveness of the proposed conceptual framework, getting insights to ease the use of blockchain query engines and for validating the proposed draft ontology. Second, providing a roadmap for the design and building of query engine tools for developers and software architects of the Polkadot ecosystem, considering use cases for retrieving and analyzing blockchain data from the Polkadot network and its parachains. These use cases encompass queries supported by a controlled natural language (CNL) and the POnto ontology.

The designed roadmap is composed by four main steps:

1. Creation of a conceptual framework;
2. Knowledge base construction: script to extract data from the polkadot ecosystem and align the data as individuals of the POnto ontology;
3. Specification of a controlled natural language (CNL) that will use the POnto ontology;
4. Query engine creation that supports users specifying queries using the CNL.

We hope this roadmap inspires developers and software architects interested in the Polkadot ecosystem expansion, by enhancing community engagement and facilitating the development of blockchain-based applications.

The selection of the participants targeted experts, developers, and general users from the Polkadot community interested in retrieving and analyzing blockchain data, including cross-information among different chains. We tried to keep the participant selection process as balanced as possible, based on the following criteria:

A. Participants' level of knowledge and experience with the Polkadot ecosystem ;
B. Their experience in retrieving and analyzing blockchain data, and;
C. Their level of familiarity with tools for retrieving blockchain data in the ecosystem.

This strategy ensured a diverse representation of expertise and experience levels within the Polkadot community.

To collect and assess relevant perspectives within the community, we elaborated an online questionnaire to gather information around the usefulness of a prospective query engine based on the POnto ontology. We chose to use Google Forms to support the creation and dissemination of this questionnaire. In addition, we conducted a collaborative session to support discussions around the proposed ontology and envisioned query engine. This online collaboration was supported by Mural with activities for conceptual mapping of the Polkadot ecosystem and query engine requirements prioritization. Both the questionnaire and Mural charts supported validating the conceptual framework (see Section VI).

The questionnaire was pilot-tested with a small team of Polkadot experts. They suggested a number of modifications concerning its scope. Initially, we tried to cover a broader perspective in terms of query ability. The idea was to consider similarities and discrepancies that should be addressed when applying the proposed conceptual framework on different domains. Afterwards, we adjusted the questionnaire and focused specifically on aspects of the Polkadot ecosystem.

In terms of data collection, we gathered quantitative and qualitative data both from the questionnaire and the collaborative session. In the current stage, the questionnaire was pilot-tested and is ready to be disseminated to collect data in the wild. We gathered the preliminary data from the pilot-test, along with the data from discussion with experts. We then analyzed the expert feedback from the questionnaire and the collaborative session to identify and categorize relevant concepts and their relationships to adjust the ontology modeling. In addition, to establish a roadmap for a future query engine.

## V. THE POnto Ontology

POnto[1] is an open-source ontology licensed under Apache 2.0. A draft version of it was initially based on Polkadot's whitepaper [10]. Afterwards, POnto was subsequently enriched according to discussions with domain experts and latest documentation available[2].

As part of a conceptual framework, POnto was designed as a modular ontology organized hierarchically and provides a representation of different aspects and features of the Polkadot ecosystem. POnto is a structured representation of the ecosystem's fundamental components, concepts, and relationships. Table 1 presents the ontology key metrics.

TABLE I. Ontology Metrics

| Metrics | |
|---|---|
| Triples | 980 |
| Classes | 140 |
| Properties | 57 |
| Object Properties | 40 |
| Datatype Properties | 5 |
| Individuals | 26 |

POnto's focus is to support developers, researchers, and enthusiasts to enhance data analysis, communicability and domain knowledge sharing within the Polkadot community. Fig. 7 illustrates the POnto ontology modeling process in the Protégé tool[3].

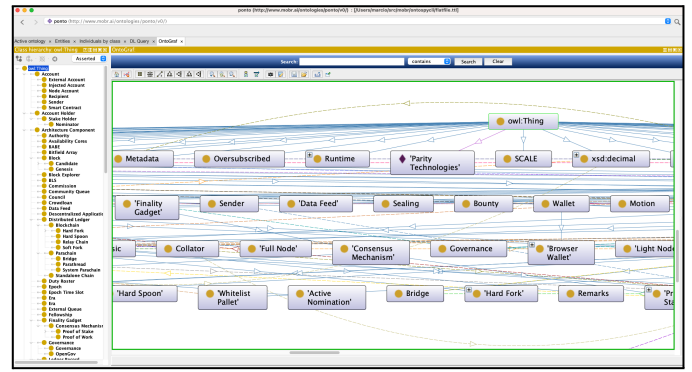

Fig. 7. POnto modeling on the Protégé tool.

The ontology currently covers eleven modules, including the Core module, and others to represent specific aspects of the Polkadot ecosystem. These other modules are Consensus, Governance, Ledger, Network, Oracle, Stake, Token, Tooling, Transaction, and Wallet.

---

[1] https://github.com/mobr-ai/POnto
[2] https://wiki.polkadot.network/
[3] https://protege.stanford.edu

POnto's *Core* module defines classes to represent common entities that serve as a sort of upper ontology, supporting the other modules. It covers entities such as architecture components, protocols, wallets, modules, pallets, decentralized applications, etc.

The *Consensus* module focuses on the mechanisms and protocols used in the ecosystem to achieve agreement and validity of transactions across multiple chains. It includes concepts such as block production, finality, consensus algorithms, and block authorship.

The governance mechanisms and processes specifications are part of the *Governance* module. It includes concepts related to on-chain governance, where voting and delegation are made through blockchain-based mechanisms. This module encompasses entities like voting systems, proposal submission, committee structures, and referendum processes.

The *Network* module focuses on the network infrastructure and protocols used in the Polkadot ecosystem. It includes concepts such as network nodes, connectivity, network synchronization, and message propagation.

The integration of external data into the Polkadot ecosystem is part of the *Oracle* module. It includes concepts related to how oracles provide external data to smart contracts and decentralized applications (dApps). The module covers data feeds, queries, and the interaction between oracles and the blockchain.

The *Stake* module describes staking mechanisms of the ecosystem. It covers concepts such as stake, stake holder accounts, and stake pools (groups of stake holders who combine their stakes).

Digital tokens and assets are covered in the *Account* module, describing concepts related to different types of account such as multisig account and proxy account, as well as concepts related to fungible and non-fungible tokens.

The *Transaction* module focuses on the representation and processing of transactions. It includes concepts such as extrinsics (representing transaction data), senders and recipients of transactions, transaction properties (e.g., amount transferred, asset type), and the association of transactions with ledger records and blocks.

The other three modules (*Ledger*, *Tooling*, and *Wallet*) aim at representing key individuals of the ecosystem to exemplify the use of the ontology. The *Ledger* module specifies the Polkadot and Kusama relay chains and some of their parachains, like Acala and Moonbeam for Polkadot and Statemine for Kusama. The *Tooling* module encompasses individuals to represent various tools of the ecosystem, including libraries, and development frameworks available for building applications. Individuals representing essential wallets of the ecosystem are part of the *Wallet* module.

## VI. VALIDATION

The validation of the conceptual framework was based on domain expert discussions (supported by a collaborative online session using Mural) and a questionnaire to collect their feedback (see Section VI.A).

We explored use cases and queries specified using a controlled natural language (CNL) based on the POnto ontology. For that end, we engaged with a team of Polkadot experts. The goal was to gather their insights and requirements for the proposed conceptual framework. In addition, we observed their perspectives about an envisioned query engine using the proposed ontology.

The outcome of the discussion with domain experts led to competency questions that were used to validate the proposed ontology, and provided insights that supported the creation of the roadmap (see Section IV). We asked domain experts to assess the usefulness of a number of queries specified in a controlled natural language. In addition, we asked experts to provide any further queries they deemed useful for retrieving information in the Polkadot ecosystem.

### A. Mural and Questionnaire

A questionnaire[4] was designed to understand the profile of users interested in such a query engine for the Polkadot ecosystem and for Polkadot blockchain data in general, and to identify common use cases and requirements for the proposed framework. Besides collecting participants' perspectives and query suggestions, this questionnaire aimed at gathering feedback on the framework and its development.

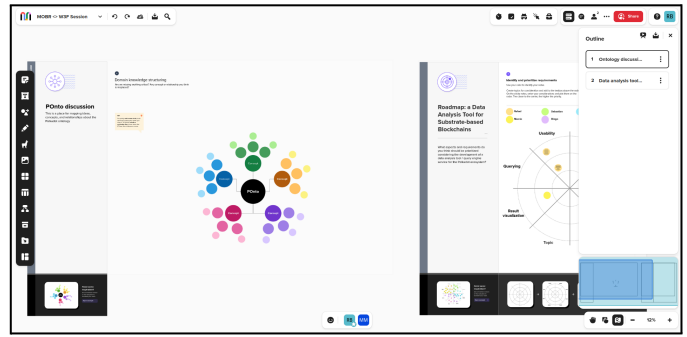

Fig. 8. Conceptual mapping and prioritization activities in Mural.

To engage with domain experts, we carried a collaborative online session using Mural[5]. After a presentation addressing the research initiative status, two activities were carried to enable engagement. First, we created a conceptual mapping practice to stimulate participants to provide their feedback over the current POnto modeling (left side of Fig. 8). Participants were asked to assess existent entities and create missing concepts and gaps in relationships. Then, in a second activity, experts were asked to give their input regarding requirements prioritization over the envisioned query engine (supported by the POnto ontology). This was carried through a radar plot where participants placed sticky notes on top of different dimensions, which they could specify (illustrated on right side of Fig. 8).

### B. Competency Questions

Competency questions (CQs) are fundamental queries that users want an ontology to support answering. Generally, they guide ontology development, validate completeness and correctness, and ensure the ontology aligns with user needs in specific domains. We structured each competency question (CQ1 to CQ6) using a controlled vocabulary (in natural language) and the respective SPARQL query. As shown in Listing 1, all competency questions use the following prefixes: *rdf*, *rdfs*, *owl*, *ponto*, and *xsd*.

---
[4] https://forms.gle/DBPXCTJRtLoigXGz6
[5] https://www.mural.co/

```
PREFIX rdf: <http://www.w3.org/1999/02/22-rdf-syntax-ns#>
PREFIX rdfs: <http://www.w3.org/2000/01/rdf-schema#>
PREFIX owl: <http://www.w3.org/2002/07/owl#>
PREFIX ponto: <http://www.mobr.ai/ontologies/ponto#>
PREFIX xsd:  <http://www.w3.org/2001/XMLSchema#>
```

Listing 1. Prefixes used throughout the competency questions.

The first competency question (CQ1) illustrates how the conceptual framework is capable of answering "what is" type of questions for any concept defined in the ontology.

**CQ1**. What is an XCM Channel?

```
# What is an XCM Channel?
SELECT ?def
WHERE {
  ponto:XCMChannel rdfs:comment ?def .
}
```

Listing 2. CQ1: What is an XCM Channel?

CQ2 focuses on how the conceptual framework supports queries related to the properties defined in the ontology.

```
# Which components are part of the polkadot architecture?
SELECT ?component WHERE {
    ponto:PolkadotArchitecture ponto:hasComponent ?component
}
```

Listing 3. CQ2: What is an XCM Channel?

CQ3 and CQ4 show how the conceptual framework may create powerful analysis that is not possible using existent block explorer applications.

```
# How many accounts have DOT and KSM tokens?
SELECT (COUNT(DISTINCT ?account) AS ?numOfAccounts)
WHERE {
  ?account ponto:hasToken ?tokenD .
  ?tokenD a ponto:DOT .
  ?account ponto:hasToken ?tokenK .
  ?tokenK a ponto:KSM .
}
```

Listing 4. CQ3: How many accounts have DOT and KSM tokens?

Regarding CQ4, it also shows how the framework could be used to correlate different components from the ecosystem. This competency question addresses the user's current account balance. CQ4 assumes there is a software component capable of retrieving the user account balance for the DOT token, which in the example is 500.1 DOTs.

```
# Which accounts have more DOT tokens than me?
SELECT ?account
WHERE {
  ?account ponto:hasToken ?token .
  ?token a ponto:DOT .
  ?token ponto:hasBalance ?balance .
  FILTER (xsd:decimal(?balance) > 500.1)
}
```

Listing 5. CQ4: Which accounts have more DOT tokens than me?

CQ5 and CQ6 validate how the framework may be applied to answer queries that integrate different sources within the ecosystem, enabling a holistic analysis.

```
# How many transactions happened between July 4th and 8th specifically in the Moonbeam parachain?
SELECT (COUNT(?transaction) AS ?transactionCount)
WHERE {
  ?transaction a ponto:Transaction ;
      ponto:timestamp ?timestamp .
  ?transaction ponto:registeredOn ?block .
  ?block ponto:composes ponto:Moonbeam .
  FILTER(?timestamp > "2023-07-04T00:00:00Z"^^xsd:dateTimeStamp && ?timestamp < "2023-07-08T00:00:00Z"^^xsd:dateTimeStamp)
}
```

Listing 6. CQ5: How many transactions happened between July 4th 2023 and July 8th 2023 specifically in the Moonbeam parachain?

```
#What are the top 5 parachains by pull requests in the last 7 days?
SELECT ?parachain
       (COUNT(?pr) as ?prCount)
WHERE {
  ?parachain a ponto:Parachain .
  ?parachain ponto:hasPR ?pr .
  ?pr ponto:opened ?openedDate .
  FILTER (?openedDate >= now() - xsd:duration('P7D'))
}
GROUP BY ?parachain
ORDER BY DESC(?prCount)
LIMIT 5
```

Listing 7. CQ6: What are the top 5 parachains by pull requests in the last 7 days?

VII. FINAL REMARKS AND FUTURE WORK

The Polkadot ecosystem poses unique challenges in terms of integrated data analysis and communicability. To address these challenges, we proposed a conceptual framework that leverages semantic web technologies to improve understanding of domain knowledge, accessibility, and analysis of ecosystem data.

The key contribution of this work is the POnto ontology, a structured representation of the Polkadot ecosystem's concepts and relationships. POnto provides a foundation for capturing and organizing domain knowledge, enabling a standardized and formal representation of the ecosystem.

To validate the proposed conceptual framework, we engaged with domain experts from the Polkadot community. Through collaborative activities and a questionnaire, we gathered valuable feedback and insights, which supported refining the ontology and outlining a roadmap for a query engine based on the POnto ontology. The experts' input confirmed the potential of the knowledge-oriented approach in supporting holistic analyses and facilitating the development of knowledge-oriented applications within the Polkadot ecosystem.

The presented study tackles a gap in the Polkadot literature, as existing works mainly focus on either technical or economic aspects. By formalizing domain knowledge through POnto, and supporting querying based on controlled natural language (CNL), we provide users, developers, and stakeholders with a clearer understanding of the ecosystem, enhancing communication, integration, and community engagement.

The POnto ontology opens up possibilities for cross-chain data analysis and integration with other functionalities of the ecosystem, enabling developers to write consistent and efficient querying code across the entire network. Additionally, it facilitates the retrieval of relevant information and fosters the exchange of domain knowledge among participants, contributing to the growth and adoption of the Polkadot ecosystem in diverse socio-technical environments.

As we move forward, the outlined roadmap will guide the development of a query engine that supports users in specifying queries using CNL and the POnto ontology. This engine will enable users to interact with the Polkadot ecosystem in an intuitive and assisted manner, promoting wider participation and fostering innovation within the community.

We believe the presented work holds significant promise for enhancing the Polkadot ecosystem. We hope it inspires further research and development of knowledge-oriented approaches for Web3, paving the way for more efficient, accessible, and inclusive decentralized technologies in the future.

ACKNOWLEDGMENT

This work was supported by a research grant from the Web3 Foundation. The authors would like to express their gratitude to the members of the Polkadot community who participated in the validation process of the proposed conceptual framework and the POnto ontology. Their valuable feedback and insights significantly contributed to the refinement of the work. We also extend our thanks to the anonymous questionnaire respondents for their thoughtful comments and suggestions that improved the overall quality of POnto and the proposed approach.

REFERENCES

[1] Cano-Benito, M., & Mera, E. (2017). A Semantically Enabled Service-Oriented Blockchain. In 2017 IEEE 9th International Conference on Semantic Computing (ICSC) (pp. 113-116). IEEE.

[2] Velasco, G. (2019). Blockchain Ontologies: The Politics of Order in Distributed Ledger Technologies. 2019 IEEE/ACM 2nd International Workshop on Emerging Trends in Software Engineering for Blockchain (WETSEB), 2-8.

[3] Janowicz, K., Przybyłek, A. S., & Scheider, S. (2019). A semantic blockchain to enhance trust in self-governed distributed systems. International Journal on Semantic Web and Information Systems (IJSWIS), 15(1), 1-22.

[4] Scrocca, M., Mariani, S., & Simi, M. (2020). Smart Contract and Semantic Web Technologies for Efficient Contract Management in the Transportation Industry. In 2020 IEEE International Conference on Blockchain and Cryptocurrency (ICBC) (pp. 17-24). IEEE.

[5] Ding, J., De Meester, B., & Verborgh, R. (2021). Towards Semantic Web Technologies to Enable Decentralized Scientific Collaboration. Proceedings of the 18th International Conference on Ontologies, DataBases, and Applications of Semantics (ODBASE '21), 30-39.

[6] Palma, G., Trojahn, C., & Hilaire, V. (2011). Ontologies in the World Wide Web. Semantic Web Technologies, 87-125.

[7] EthOn - Ethereum Ontology. (n.d.). Retrieved from https://github.com/dlt-erc20/ethon

[8] BLONDIE - BLockchain ONtology with Dynamic Extensibility. (n.d.). Retrieved from https://blondie-ontology.github.io/blondie/

[9] Distel, F., Ahmetaj, S., Thimm, M., & Auer, S. (2019). The DLT Ontology: A Modular Ontology for the Description of Distributed Ledger Systems. In Proceedings of the Joint Ontology Workshops 2019 (JOWO 2019) (Vol. 2670).

[10] Polkadot whitepaper – POLKADOT: VISION FOR A HETEROGENEOUS MULTI-CHAIN FRAMEWORK (n.d.) Retrieved from https://assets.polkadot.network/Polkadot-whitepaper.pdf

[11] Yin, R. K. (2009). Case study research: design and methods (4th ed). Los Angeles, Calif: Sage Publications.